\documentclass[sigconf]{acmart}

\copyrightyear{2026}
\acmYear{2026}
\setcopyright{cc}
\setcctype{by}
\acmConference[WWW '26]{Proceedings of the ACM Web Conference 2026}{April 13--17, 2026}{Dubai, United Arab Emirates}
\acmBooktitle{Proceedings of the ACM Web Conference 2026 (WWW '26), April 13--17, 2026, Dubai, United Arab Emirates}
\acmPrice{}
\acmDOI{10.1145/3774904.3792143}
\acmISBN{979-8-4007-2307-0/2026/04}

\usepackage{booktabs}
\usepackage{multirow}
\usepackage{makecell}
\usepackage{threeparttable}
\usepackage[table]{xcolor}
\usepackage{pifont} 
\usepackage{soul}
\usepackage{subcaption}
\usepackage[most]{tcolorbox}
\tcbset{colback=gray!5!white, colframe=black!75!black, boxrule=0.5pt, arc=2pt}
\usepackage{enumitem}
\usepackage{geometry}
\tcbuselibrary{listingsutf8}

\definecolor{gray1}{RGB}{242, 242, 242}
\definecolor{gray2}{RGB}{225, 225, 225}
\definecolor{gray3}{RGB}{205, 205, 205}
\definecolor{gray4}{RGB}{191, 191, 191}

\AtBeginDocument{%
  }

\begin{document}

\title{ES-MemEval: Benchmarking Conversational Agents on Personalized Long-Term Emotional Support}



\settopmatter{authorsperrow=4}

\author{Tiantian Chen}
\email{2111287@tongji.edu.cn}
\orcid{0000-0001-7346-7670}
\affiliation{%
  \institution{Tongji University}
  \city{Shanghai}
  \country{China}
}

\author{Jiaqi Lu}
\email{2512091@tongji.edu.cn}
\orcid{0009-0006-9236-4531}
\affiliation{%
  \institution{Tongji University}
  \city{Shanghai}
  \country{China}
}

\author{Ying Shen}
\authornote{Ying Shen is the corresponding author.}
\email{yingshen@tongji.edu.cn}
\orcid{0000-0002-2966-7955}
\affiliation{%
  \institution{Tongji University}
  \city{Shanghai}
  \country{China}
}

\author{Lin Zhang}
\email{cslinzhang@tongji.edu.cn}
\orcid{0000-0002-4360-5523}
\affiliation{%
  \institution{Tongji University}
  \city{Shanghai}
  \country{China}
}

\renewcommand{\shortauthors}{Tiantian Chen, Jiaqi Lu, Ying Shen, and Lin Zhang}

\begin{abstract}

Large Language Models (LLMs) have shown strong potential as conversational agents. Yet, their effectiveness remains limited by deficiencies in robust long-term memory—particularly in complex, long-term Web-based services such as online emotional support. However, existing long-term dialogue benchmarks primarily focus on static and explicit fact retrieval, failing to evaluate agents in these critical scenarios where user information is dispersed, implicit, and continuously evolving. 
To address this gap, we introduce ES-MemEval, a comprehensive benchmark that systematically evaluates five core memory capabilities—information extraction, temporal reasoning, conflict detection, abstention, and user modeling—in long-term emotional support scenarios, covering question answering, summarization, and dialogue generation tasks. To support the benchmark, we also propose EvoEmo, the first multi-session dataset for personalized long-term emotional support scenarios, capturing fragmented, implicit user disclosures and evolving user states. Extensive experiments on open-source long-context, commercial, and retrieval-augmented (RAG) LLMs reveal that explicit long-term memory is essential to reduce hallucinations and enable effective personalization. At the same time, RAG enhances factual consistency but struggles with temporal dynamics and evolving user states. These findings highlight both the potential and limitations of current paradigms, encouraging the development of more robust memory–retrieval integration in long-term personalized dialogue systems. 
\end{abstract}

\ccsdesc[500]{Information systems~Personalization}
\ccsdesc[500]{Human-centered computing~User models}
\ccsdesc[500]{Computing methodologies~Natural language generation}

\keywords{long-term dialogue; emotional support; personalization; conversational agents; large language models; user modeling}


\maketitle

\newcommand\webconfavailabilityurl{https://doi.org/10.5281/zenodo.18338564}
\ifdefempty{\webconfavailabilityurl}{}{
\begingroup\small\noindent\raggedright\textbf{Resource Availability:}\\
The benchmark dataset and source code for this paper are publicly available via Zenodo at
\url{\webconfavailabilityurl}, with the source code hosted at
\url{https://github.com/slptongji/ES-MemEval}.
\endgroup
}

\begin{figure}[bh]
  \centering
  \includegraphics[width=0.85\linewidth]{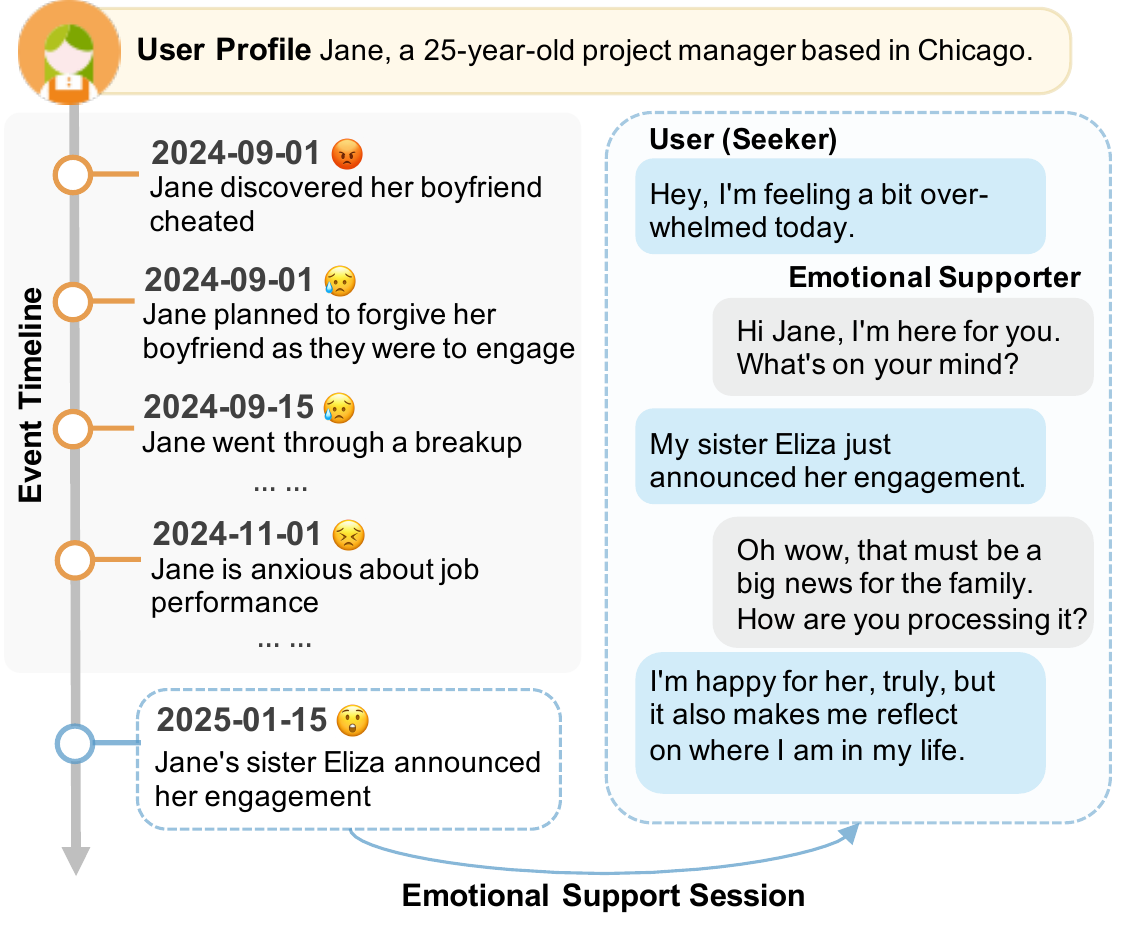}
  \caption{
Excerpt from the EvoEmo dataset, illustrating fragmented disclosures over months. 
Comprehending why the \textit{sister’s engagement} evokes \textit{overwhelm} requires recalling the \textit{earlier breakup}, emphasizing the importance of robust long-term memory in emotional support dialogues.
  }
  \Description{A conversation example is shown.}
  \label{fig:esc_example} 
\end{figure}

\section{Introduction}


\begin{table*}[ht]
\centering
\renewcommand{\arraystretch}{0.95}
\caption{
Comparison of long-term dialogue benchmarks by dataset scale, coverage of core memory abilities, and evaluation objectives. 
Statistics include total long-term conversations (\textit{Tot. Conv.}), average sessions per conversation (\textit{Avg. Sess.}), and average turns per session (\textit{Avg. Turn.}). 
Core memory abilities are abbreviated as \textit{IE} (Information Extraction), \textit{TR} (Temporal Reasoning), \textit{CD} (Conflict Detection), \textit{Abs} (Abstention), and \textit{UM} (User Modeling).
}
\resizebox{0.85\linewidth}{!}{
\begin{tabular}{ccccccccc>{\centering\arraybackslash}p{6cm}}
\toprule
\multirow{2}{*}{\textbf{Benchmark}}
& \multicolumn{3}{c}{\textbf{Statistics}} 
& \multicolumn{5}{c}{\textbf{Core Memory Abilities}} 
& \multirow{2}{*}{\textbf{Overall Goal}} \\
\cmidrule(lr){2-4} \cmidrule(lr){5-9}
& \textbf{Tot. Conv.} 
& \textbf{Avg. Sess.} 
& \textbf{Avg. Turn.} 
& \textbf{IE} 
& \textbf{TR} 
& \textbf{CD} 
& \textbf{Abs} 
& \textbf{UM} 
& \\
\midrule
MSC \cite{r-1}             & 5K   & 3.4   & 12.6     & \textcolor{red}{\ding{55}}  &  \textcolor{red}{\ding{55}}  &  \textcolor{red}{\ding{55}}  &  \textcolor{red}{\ding{55}} 
&  \textcolor{red}{\ding{55}}   & Open-domain dyadic chit-chat \\

Conversation Chronicles \cite{r-10}             & 200K   & 5   & 11.7     & \textcolor{red}{\ding{55}}  &  \textcolor{red}{\ding{55}}  &  \textcolor{red}{\ding{55}}  &  \textcolor{red}{\ding{55}} 
&  \textcolor{red}{\ding{55}}   & Open-domain dyadic chit-chat \\

DuLeMon \cite{r-3}        & -    & 24.5K   & 16.3     & \textcolor{red}{\ding{55}} & \textcolor{red}{\ding{55}} & \textcolor{red}{\ding{55}}   &   \textcolor{red}{\ding{55}} &  \textcolor{green}{\ding{51}}  & Personalized open-domain conversation \\

MemoryBank \cite{intro-3}     & 10   & 15     & 7.6      & \textcolor{green}{\ding{51}}& \textcolor{green}{\ding{51}}& \textcolor{red}{\ding{55}}   &   \textcolor{red}{\ding{55}} &  \textcolor{red}{\ding{55}}  & Personalized conversational QA \\

PerLTQA  \cite{intro-6}       & 141  & 21.3   & 8.4      & \textcolor{green}{\ding{51}}& \textcolor{green}{\ding{51}}& \textcolor{red}{\ding{55}}   &   \textcolor{red}{\ding{55}} & \textcolor{green}{\ding{51}} & Personalized conversational QA \\

LOCOMO \cite{intro-7}         & 10   & 27.2   & 21.6     & \textcolor{green}{\ding{51}}& \textcolor{green}{\ding{51}}& \textcolor{red}{\ding{55}}& \textcolor{green}{\ding{51}} & \textcolor{red}{\ding{55}}& Open-domain dyadic chit-chat \\

LongMemEval \cite{intro-5}    & -    & 50K   & -  & \textcolor{green}{\ding{51}}& \textcolor{green}{\ding{51}}& \textcolor{red}{\ding{55}} & \textcolor{green}{\ding{51}}& \textcolor{red}{\ding{55}}& Factual and behavioral assistant QA \\

MADial-Bench \cite{r-4}     & 2    & 80   & 9.2  & \textcolor{green}{\ding{51}}& \textcolor{red}{\ding{55}}& \textcolor{red}{\ding{55}}& \textcolor{red}{\ding{55}}& \textcolor{red}{\ding{55}}& Child–assistant emotional dialogue \\

DialSim  \cite{r-2}         & 3    & 1.3K   &   -  & \textcolor{green}{\ding{51}}     & \textcolor{green}{\ding{51}}&  \textcolor{red}{\ding{55}}  &   \textcolor{green}{\ding{51}}  &  \textcolor{red}{\ding{55}}  &  TV-script-based multi-party chit-chat \\

\textbf{ES-MemEval} & \textbf{18} & \textbf{22.3} & \textbf{23.4} & \textcolor{green}{\ding{51}}& \textcolor{green}{\ding{51}}& \textcolor{green}{\ding{51}}& \textcolor{green}{\ding{51}}& \textcolor{green}{\ding{51}}& Personalized emotional support conversation \\
\bottomrule
\end{tabular}
}
\label{tab:benchmark_comparison}

\end{table*}

\begin{figure*}[thb]  
  \centering
\includegraphics[width=0.8\linewidth]{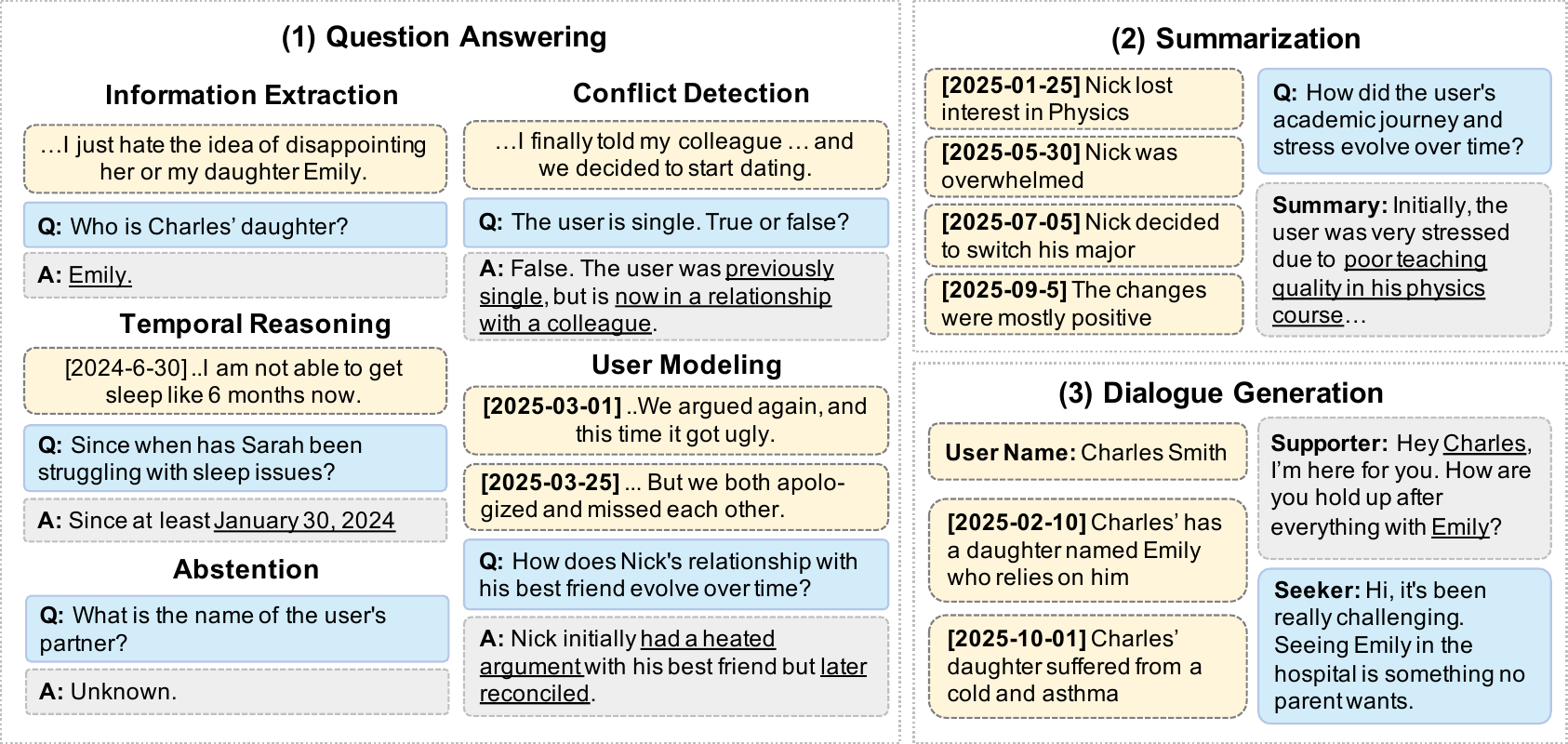}
  \caption{Overview of ES-MemEval, comprising three tasks—QA, summarization, and dialogue generation—designed to evaluate five core capabilities critical for long-term personalized dialogue agents.}
  \label{fig:benchmark} 
\Description{Overview of ES-MemEval. The framework includes three tasks (QA, summarization, dialogue generation) designed to evaluate five core capabilities essential for long-term personalized dialogue agents.}
\end{figure*}


Large language models (LLMs) have demonstrated remarkable potential as conversational agents, enabling widespread deployment across Web platforms for applications such as customer support and mental health services \cite{intro-2, intro-1, intro-10}. 
While they excel in short-term exchanges, their effectiveness remains limited in complex, long-term scenarios such as online emotional support (ES), which require agents to track evolving user states and integrate implicit, fragmented user disclosures across multiple sessions \cite{intro-3, intro-4, intro-13}. 
As illustrated in Figure~\ref{fig:esc_example}, robust long-term memory is therefore essential—not only to generate personalized and coherent responses, but also to mitigate hallucinations that could undermine user trust in these sensitive Web-based services \cite{intro-11, intro-12}.

However, existing dialogue benchmarks inadequately evaluate LLMs' long-term memory ability in such \textit{implicit}, \textit{fragmented}, and \textit{evolving} contexts \cite{intro-8, intro-14, intro-15}.
Most current benchmarks narrowly focus on \textit{static and explicit fact retrieval}—for instance, recalling named entities and event details in QA-style tasks—where relevant information is explicit and largely stable over time \cite{intro-3, intro-5, intro-6}.
Consequently, they capture only a limited facet of long-term memory, overlooking the reasoning and abstraction processes essential for emotional support dialogues.
In these scenarios, agents must not only extract and comprehend dispersed user information but also summarize over evolving user states and ultimately generate personalized, contextually grounded responses across sessions.
Yet, current benchmarks lack systematic means to evaluate how models \textit{integrate}, \textit{abstract}, and \textit{apply} user information throughout extended, emotionally complex interactions.

To bridge this gap, we introduce \textbf{ES-MemEval}, the first comprehensive benchmark tailored to long-term emotional support scenarios.
Unlike prior benchmarks centered on static, explicit factual recall, ES-MemEval systematically evaluates LLMs’ ability to integrate, abstract, and apply evolving, implicit, and fragmented user information across multiple sessions.
As illustrated in Figure \ref{fig:benchmark}, the benchmark comprises three complementary tasks: \textit{question answering}, \textit{summarization}, and \textit{dialogue generation}.
The question answering task examines models' ability to retrieve and comprehend information scattered across extended interactions.
The summarization task assesses their capacity to abstract and synthesize user state dynamics over time. 
The dialogue generation task directly measures models' proficiency in effectively leveraging long-term memory to deliver personalized emotional support.
Together, these tasks provide a rigorous evaluation of five key long-term memory capabilities—\textit{information extraction}, \textit{temporal reasoning}, \textit{conflict detection}, \textit{abstention}, and \textit{user modeling}—that underpin trustworthy and personalized conversational agents.


To support the proposed benchmark, we construct \textbf{EvoEmo}, the first multi-session dataset for long-term emotional support scenarios featuring evolving user states. It contains multi-session conversations involving 18 virtual users seeking emotional support, averaging 510 turns ($\approx$13.3k tokens) across up to 33 sessions per user. EvoEmo combines sessions drawn from real emotional support data with sessions generated from detailed user profiles and temporally and causally structured event timelines, thereby realistically capturing the fragmented user disclosures and longitudinal evolution of user states. This dataset offers a reliable foundation for studying longitudinal personalization and complex user modeling in emotionally sensitive contexts.

We further conduct systematic experiments on ES-MemEval with open-source long-context, commercial, and retrieval-augmented (RAG) \cite{intro-9} LLMs across the aforementioned tasks. 
Experimental results yield several insights into long-term personalized emotional support.
\textbf{First}, without explicit histories, models tend to hallucinate user experiences, undermining reliability and personalization. 
\textbf{Second}, while RAG enhances factual consistency and alignment with user experiences, it struggles with temporal dynamics and evolving user states, highlighting the necessity for retrieval-aware calibration.
\textbf{Third}, personalization strongly correlates with long-term memory, whereas emotional support proves less memory-sensitive and often relies on general strategies.
\textbf{Fourth}, session-level retrieval best captures evolving user information but may introduce redundancy. 
\textbf{Fifth}, smaller long-context models degrade with extra-long inputs, underscoring the need to integrate retrieval with external memory mechanisms. 
\textbf{Finally}, RAG narrows the gap between open-source and commercial systems by enhancing the personalization and memory alignment of generated responses.
Collectively, these findings highlight the strengths and limitations of existing paradigms, suggesting promising directions for future research.

Our contributions can be summarized as follows:
(1) We present EvoEmo, the first multi-session dataset specifically designed for personalized long-term emotional support scenarios, capturing evolving user states and implicit, fragmented disclosures.
It provides a valuable foundation for studying longitudinal user modeling and personalization in emotionally sensitive contexts.
(2) We introduce ES-MemEval, a novel and comprehensive benchmark that evaluates five essential long-term memory capabilities of dialogue agents across three tasks in complex, long-term emotional support scenarios.
(3) We conduct extensive experiments across major LLM paradigms, yielding empirical insights into the strengths and limitations of existing paradigms and informing future research directions in long-term personalized emotional support.

\section{Related Work}
\subsection{Long-Term Dialogue Benchmarks}

Effective long-term memory and support for multi-session interactions pose fundamental challenges in conversational AI \cite{intro-7, r-1}.
To evaluate these abilities, recent benchmarks have shifted from open-ended dialogue generation tasks to QA-style evaluations that more directly test retrieval and reasoning \cite{r-11, r-12}.
\citet{intro-3} offered multi-day dialogues with 15 virtual users and 194 QA samples for cross-session retrieval. 
\citet{intro-7} provided 10 extended dialogues with questions spanning single-hop, multi-hop, temporal, commonsense, and adversarial reasoning. 
\citet{intro-6} contributed a large-scale QA benchmark of 3,409 dialogues and 8,593 questions covering world knowledge, personal profiles, social relations, and dialogue events. 
\citet{intro-5} targeted chat assistants with 500 questions testing five memory-related skills: information extraction, multi-session reasoning, temporal reasoning, knowledge updating, and abstention. 
\citet{r-4} constructed 160 simulated child–assistant dialogues and introduces a human-centered evaluation framework for proactive and passive recall.

Despite these advances, many benchmarks remain largely confined to fact-centric QA, where information is explicit and tasks are clearly defined \cite{intro-3, intro-6, intro-5}. 
They fail to capture the complexities of real interactions, specifically implicit expressions, fragmented information, and evolving user states.
Moreover, their reliance on narrow evaluation formats (e.g., QA or retrieval) constrains the assessment of cross-session reasoning and personalized response generation. 
To address these gaps, we introduce ES-MemEval, a benchmark for long-term emotional support dialogues that includes QA, summarization, and dialogue generation tasks, enabling systematic evaluation of memory utilization and personalized adaptation.
Table \ref{tab:benchmark_comparison} presents a comparative analysis of existing long-term dialogue benchmarks and datasets against ES-MemEval.

\begin{figure*}[thb]
  \centering
\includegraphics[width=0.85\linewidth]{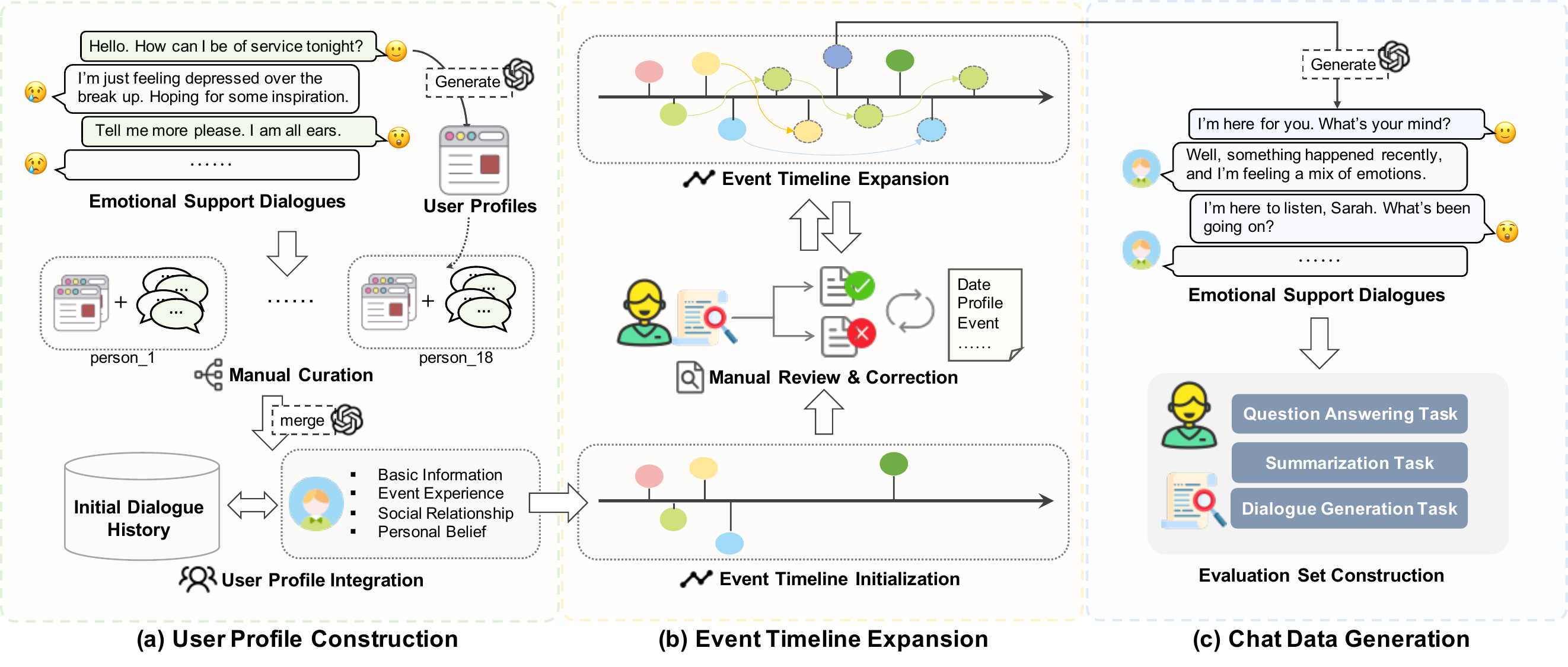}
  \caption{
The data generation pipeline of EvoEmo, consisting of three stages: (a) user profile construction, (b) event timeline expansion, and (c) chat data generation, aiming to simulate realistic long-term emotional support conversations.
  }
  \label{fig:data_contrsuction} 
\Description{The data construction process of the EvoEmo dataset.}
\end{figure*}

\subsection{Emotional Support Conversation}

Emotional support aims to alleviate emotional distress and help individuals cope with life challenges \cite{intro-8}. 
With increasing demand for mental health care and companionship, ES dialogue has become a growing research focus. 
\citet{r-5} collected \textasciitilde25k short empathetic dialogues grounded in emotional scenarios.
\citet{intro-8} introduced ESConv, a crowdsourced dataset of 1,035 multi-turn ES dialogues annotated with support strategies. 
\citet{r-6} harvested large-scale counseling dialogues from an online platform to model real consultation processes, while \citet{r-7} released PsyDial, a de-identified and publicly available version. 
Beyond datasets, other works \cite{r-8, r-6, r-9} proposed diverse ES generation methods, such as positive guidance, reflection understanding, and self-disclosure, to improve supportiveness and interaction quality.

Despite these advances, most efforts focus on limited-turn or single-session dialogues, with limited modeling of long-term user trajectories and evolving states. 
In practice, emotional support often spans days or weeks, posing higher demands on long-term memory utilization and personalized adaptation. 
To bridge this gap, we introduce EvoEmo, a dataset of long-term emotional support dialogues capturing the evolution of user states, and ES-MemEval, a benchmark for evaluating models’ ability to maintain and leverage long-term memory to support personalized adaptation.

\section{EvoEmo Dataset}



To facilitate the evaluation of personalized conversational agents in complex long-term scenarios, we propose EvoEmo, a long-term emotional support dialogue dataset that captures the dynamic evolution of user states.
The data generation pipeline consists of three stages: user profile construction, event timeline expansion, and chat data generation, as shown in Figure \ref{fig:data_contrsuction}.

\subsection{User Profile Construction}
 

To model the longitudinal evolution of user states while preserving privacy, we created 18 virtual users with meticulously designed, diverse personality traits. Each user profile was manually curated based on multiple seed dialogues sampled from the ESConv dataset \cite{intro-8}, a real-world collection of short-term emotional support conversations. The profiles were carefully expanded to include \textit{demographic information}, \textit{social relationships}, and \textit{core beliefs}, thereby providing rich and realistic representations of users’ thoughts and behavior patterns. In addition, each user’s initial dialogue history consists of these seed dialogues, which inherently reflect implicit, fragmented user disclosures and evolving user states, forming a solid foundation for subsequent multi-session dialogue generation.

\subsection{Event Timeline Expansion}


The initial sessions of each virtual user were derived from short-term ESConv dialogues of different emotional seekers, lacking long-term continuity and causal structure. 
Therefore, we constructed an event timeline for each user to simulate the longitudinal evolution of user states. 
The initial events were first generated by GPT-4o based on the initial sessions and refined by human annotators, including \textit{timestamps} and \textit{event descriptions}. 
After initialization, we employed GPT-4o to iteratively expand each event timeline in two rounds, where new events were generated conditioned on the existing event sequences and user profiles. 
Human annotators reviewed and adjusted each round to ensure temporal and causal consistency.
By extending prior events and modeling user state evolution, the resulting timelines capture both causal and experiential variation, averaging 24.8 events per user.
This iterative construction enables dynamic and causally coherent user trajectories, advancing beyond prior static user profile settings \cite{intro-3, intro-6}.

\subsection{Chat Data Generation}


Based on the constructed user profiles and event timelines, we prompted GPT-4o to generate multi-turn emotional support sessions for each user. 
GPT-4o was conditioned on structured inputs, including the current event, user profile, and summaries of relevant prior sessions, aiming to produce sessions that reflect the evolving user states and the implicit and fragmented user disclosures. 
The generated sessions were then enriched with auxiliary annotations, including \textit{emotion category}, \textit{topic}, \textit{summary}, and turn-level \textit{user observations} to facilitate downstream analyses. 
Six annotators subsequently reviewed each session for consistency with the corresponding user profile and across sessions, ensuring high data quality. 
Through this pipeline, we developed \textit{EvoEmo}, a dataset simulating 18 virtual users with evolving states and event trajectories, providing a structured and curated testbed for research on long-term emotional support and longitudinal user modeling.

\section{ES-MemEval Benchmark}

\subsection{Task Formulation}

To systematically assess dialogue agents’ long-term memory capabilities, we introduce \textit{ES-MemEval}, a benchmark comprising question answering, summarization, and dialogue generation tasks, as illustrated in Figure \ref{fig:benchmark}.

\subsubsection{Long-Term Memory Capabilities}

ES-MemEval defines five core long-term memory capabilities:
(1) \textbf{Information Extraction}— identifying key facts within and across sessions; 
(2) \textbf{Temporal Reasoning}—inferring temporal order and causal dependencies among events to track evolving user trajectories; 
(3) \textbf{Conflict Detection}—detecting and resolving contradictions in long-term memory units to maintain alignment with the user’s current state; 
(4) \textbf{Abstention}—withholding responses when available information is insufficient to ensure reliability; 
and (5) \textbf{User Modeling}—inferring and updating user traits, preferences, and states over time to enable personalized support.

\subsubsection{Evaluation Task Formats}

ES-MemEval evaluates these capabilities via three complementary tasks:
(1) \textbf{Question Answering}—assessing information retrieval and integration across sessions, spanning all five core capabilities;
(2) \textbf{Summarization}—analyzing cross-session information and user state dynamics, which focuses on temporal reasoning and user modeling;
(3) \textbf{Dialogue Generation}—simulating realistic interactions to evaluate context understanding, user modeling, and ES response generation, reflecting the holistic integration of multiple capabilities.

\subsection{Evaluation Sets \label{sec:test_set}}

To operationalize the evaluation tasks, we construct three task-specific evaluation sets, with data statistics reported in Table~\ref{tab:statistics}.

\subsubsection{Question Answering}
QA samples were generated using GPT-4o based on each virtual user’s multiple sessions and event timeline, designed to span all five core capabilities. 
Each sample includes a question type label, question text, reference answer, and supporting evidence passage.
To ensure quality, each sample was reviewed and corrected by one of six annotators, addressing issues such as type mismatches, incorrect or incomplete answers, and missing evidence. 
Following the rigorous quality assurance process, we constructed a final QA evaluation set of 1,209 high-quality samples.

\subsubsection{Summarization}
Summarization cases were constructed to evaluate agents’ ability to abstract information and perform reasoning across multiple sessions.
Specifically, GPT-4o first grouped sessions into thematic groups based on users’ event timelines. 
For each group, GPT-4o generated complex cross-session summarization questions and candidate answers, which were subsequently reviewed and refined by two annotators. 
The final evaluation set consists of 125 high-quality cases, each requiring models to extract, integrate, and summarize user states, temporal event sequences, and behavioral logic across multiple sessions.

\subsubsection{Dialogue Generation}
Dialogue scenarios were constructed to evaluate personalized dialogue generation under realistic conditions. 
GPT-4o generated candidate topics grounded in each user’s timeline and dialogue summaries. 
Each topic specification included an overview, specific details, the physical and psychological state of the user, and relevant prior sessions.
After manual review and refinement, we obtained 34 topics designed to support extended conversations about users’ evolving experiences.

\subsection{Dataset Statistics and Analysis}

Table~\ref{tab:statistics} summarizes key statistics of \textit{EvoEmo} and \textit{ES-MemEval}.
EvoEmo exhibits substantial length and complexity, with an average of 27.2 sessions and 13.3K tokens per conversation, reflecting the challenges of long-term emotional support interactions.
The QA benchmark in ES-MemEval includes 1,209 questions evenly distributed across information extraction, temporal reasoning, user modeling, conflict detection, and abstention, enabling systematic evaluation of these essential competencies. 
The summarization benchmark comprises 125 cases, with 59.2\% focused on temporal reasoning and 40.8\% on user modeling, emphasizing the challenge of cross-session integration and evolving user state tracking. 
The dialogue generation benchmark contains 34 scenarios, designed to assess models’ ability to generate personalized, long-term ES responses while coordinating the five memory abilities across multiple sessions.
Overall, these statistics highlight both the scale and balanced task design of ES-MemEval, providing a robust foundation for studying personalized long-term dialogue. Additional analyses of the dataset are provided in Appendix~\ref{app:statistics}.

\begin{table}[bt]
\centering
\caption{Statistics of the EvoEmo dataset and the derived ES-MemEval benchmark.}
\resizebox{0.8\linewidth}{!}{
\begin{tabular}{l r}
\toprule
\textbf{Conversation Statistics} & \textbf{\# Count} \\
\midrule
Avg. time span (months) / conversation & 14.9 \\
Avg. sessions / conversation & 22.3 \\
Avg. turns / session & 23.4 \\
Avg. tokens / conversation & 13,291.6 \\
Avg. tokens / session & 596.6 \\
\textbf{Total conversations} & \textbf{18} \\
\textbf{Total sessions} & \textbf{401} \\
\midrule
\textbf{QA Benchmark Statistics} & \\
\midrule
Information extraction & 271 (22.4\%) \\
Temporal reasoning & 236 (19.5\%) \\
Conflict detection & 226 (18.7\%) \\
User modeling & 251 (20.8\%) \\
Abstention & 225 (18.6\%) \\
\textbf{Total questions} & \textbf{1,209} \\
\midrule
\textbf{Summarization Benchmark Statistics} & \\
\midrule
Temporal reasoning & 74 (59.2\%) \\
User modeling & 51 (40.8\%) \\
\textbf{Total summaries} & \textbf{125} \\
\midrule
\textbf{Dialogue Generation Benchmark Statistics} & \\
\midrule
Avg. turns / session  & 20 \\
\textbf{Total scenarios} & \textbf{34} \\
\bottomrule
\end{tabular}
}
\label{tab:statistics}
\end{table}

\subsection{Evaluation Protocols}

We design the following protocols to evaluate model performance across the three task formats.

\subsubsection{Question Answering}

Answer quality is assessed using F1-Score \cite{m-6}, BERTScore \cite{m-7}, and LLM-as-Judge \cite{m-8}. 
\textit{F1-Score} measures token-level overlap with the reference answer, while \textit{BERTScore} computes semantic similarity through contextual embeddings. 
In addition, \textit{LLM-as-Judge} enables flexible evaluation of model responses: GPT-4o receives the question, reference answer, and model response, and assigns a score of 0, 1, or 2 reflecting semantic consistency. 
The detailed prompt design is described in Appendix \ref{sec:app_metric}. 
To further evaluate memory recall, we also report \textit{Recall@k} \cite{m-11} and \textit{nDCG@k} \cite{m-12} for retrieval accuracy.

\subsubsection{Summarization}

Summarization quality is evaluated using ROUGE \cite{m-9}, LLM-as-Judge, and event-based metrics inspired by FActScore \cite{m-10}. 
\textit{ROUGE-1}, \textit{ROUGE-2}, and \textit{ROUGE-L} measure lexical overlap with reference summaries at unigram, bigram, and longest common subsequence levels.
\textit{LLM-as-Judge} enables semantic evaluation of summary quality: GPT-4o is prompted with the reference summary and the model-generated summary, and assigns a score from 0 to 5 evaluating semantic consistency and faithfulness (see Appendix \ref{sec:app_metric} for details).
To further assess factual coverage, we design \textit{event-based metrics}.
GPT-4o extracts discrete events from both the reference and generated summaries, and Recall, Precision, and F1 are computed to assess alignment between these event sets.

\subsubsection{Dialogue Generation}

During evaluation, GPT-4o acted as the simulated user and generated coherent inputs from predefined scenarios to drive interactive sessions. 
To assess models under these open-ended and personalized conditions, we employed two LLM-based protocols: \textit{observation-based metrics} and \textit{LLM rating metrics}. 
The former assesses whether system responses accurately reflect user states and experiences, as captured by observation annotations from scenario-related sessions, and reports recall and weighted accuracy scores (details in Appendix~\ref{sec:app_metric}). 
The latter uses GPT-4o to rate overall dialogue quality on a 5-point scale across \textit{long-term memory}, \textit{personalization}, and \textit{emotional support}, with the abbreviated prompt provided in Appendix~\ref{sec:app_metric}. 
Together, these protocols provide a rigorous assessment of a model’s ability to integrate long-term memory with user information.

\section{Experimental Setup}

Experiments on ES-MemEval evaluate three LLM paradigms: open-source long-context models, commercial models, and their retrieval-augmented variants. 
For open-source long-context models, we select three widely adopted representatives, each supporting a 128K-token context length: Ministral-8B-Instruct-2410 \cite{m-2}, Phi-3-Medium-128k-Instruct \cite{m-3}, and Mistral-Small-3.1-24B-Instruct-2503 \cite{m-4}. 
For commercial models, we include gpt-3.5-turbo \cite{m-5} with a 4K context window and gpt-4o \cite{m-1} with a 16K context window to provide mid-range closed-source baselines. 
In addition, we set up retrieval-augmented configurations for all five models, in which a dense retriever (bge-m3 \cite{m-13}) retrieves the top-4 most relevant full-session contexts from a FAISS index \cite{m-14} to supply user information.
These setups enable a systematic comparison of open-source long-context, commercial, and retrieval-augmented paradigms with the unified ES-MemEval framework, highlighting their relative strengths and limitations in long-term personalized emotional support.
All experiments were conducted on an A100 GPU equipped with 80GB of memory.

\begin{table*}[ht]
\centering
\renewcommand{\arraystretch}{1.15}
\setlength{\tabcolsep}{4.5pt}
\caption{Performance on ES-MemEval QA task across key capabilities: Information Extraction (IE), Temporal Reasoning (TR), Conflict Detection (CD), Abstention (Abs), and User Modeling (UM). Higher scores indicate better performance.}
\resizebox{0.95\linewidth}{!}{
\begin{tabular}{cccccccccccccccccccc}
\toprule
\multirow{2}{*}{\textbf{Category}} & \multirow{2}{*}{\textbf{Model}} 
& \multicolumn{6}{c}{\textbf{F1 Score (\%) } $\uparrow$} 
& \multicolumn{6}{c}{\textbf{BERTScore (\%) } $\uparrow$} 
& \multicolumn{6}{c}{\textbf{LLM-as-Judge (0-2)} $\uparrow$} \\
\cmidrule(r){3-8} \cmidrule(lr){9-14} \cmidrule(l){15-20}
 &  
 & IE & TR & CD & Abs & UM & All 
 & IE & TR & CD & Abs & UM & All 
 & IE & TR & CD & Abs & UM & All \\
\midrule
\multirow{3}{*}{\textbf{Base (128K)}} 
& Mistral-8B           & 3.2 & 9.3 & 6.3 & 0.4 & 10.6 & 5.7 & 32.7 & 43.8 & 42.8 & 26.4 & 49.0 & 38.4 & 0.33 & 0.40 & 0.36 & 0.43 & 0.59 & 0.42 \\
& Phi-3-Medium         & 11.0 & 12.8 & 14.0 & 0.0 & \textbf{11.5} & 9.7 & 42.2 & 53.1 & 51.2 & 26.3 & \textbf{55.0} & 45.0 & 0.94 & 0.68 & 0.89 & 0.63 & 0.78 & 0.79 \\
& Mistral-24B         & \textbf{13.4} & \textbf{18.0} & \textbf{20.1} & \textbf{15.6} & 10.9 & \textbf{15.5} & \textbf{42.3} & \textbf{53.4} & \textbf{54.2} & \textbf{37.0} & 52.7 & \textbf{47.4} & \textbf{1.03} & \textbf{0.84} & \textbf{0.96} & \textbf{1.20} & \textbf{0.96} & \textbf{1.01} \\
\midrule
\multirow{3}{*}{\textbf{Base + RAG}} 
& Mistral-8B + RAG     & 8.4 & 13.3 & 18.4 & 0.2 & 12.6 & 10.3 & 39.7 & 50.1 & 52.3 & 28.1 & 51.5 & 43.8 & 0.94 & 0.76 & 1.11 & 0.63 & 0.78 & 0.85 \\
& Phi-3-Medium + RAG   & 13.2 & 16.8 & 18.7 & 0.0 & 13.4 & 12.2 & 43.3 & \textbf{54.9} & 55.1 & 27.0 & 56.1 & 46.6 & 1.21 & 0.68 & 1.25 & 0.73 & 1.04 & 0.99 \\
& Mistral-24B + RAG    & \textbf{26.7} & \textbf{18.4} & \textbf{20.4} & \textbf{11.1} & \textbf{16.4} & \textbf{18.8} & \textbf{50.4} & 53.9 & \textbf{55.5} & \textbf{36.2} & \textbf{57.5} & \textbf{50.4} & \textbf{1.42} & \textbf{1.04} & \textbf{1.32} & \textbf{1.43} & \textbf{1.07} & \textbf{1.27} \\
\midrule
\multirow{2}{*}{\textbf{Commercial}} 
& GPT-3.5-turbo(4K)                 & 10.6 & \textbf{22.4} & \textbf{21.4} & 1.1 & 17.2 & 14.0 & 41.9 & 56.4 & \textbf{54.1} & 30.1 & 58.0 & 47.4 & 0.58 & 0.60 & 0.89 & 0.80 & 0.82 & 0.73 \\
& GPT-4o(16K)                       & \textbf{20.2} & 19.6 & 12.1 & \textbf{66.7} & \textbf{21.7} & \textbf{26.6} & \textbf{48.3} & \textbf{57.1} & 49.9 & \textbf{57.0} & \textbf{60.4} & \textbf{54.2} & \textbf{1.22} & \textbf{1.13} & \textbf{1.00} & \textbf{1.67} & \textbf{1.13} & \textbf{1.25} \\
\midrule
\multirow{2}{*}{\textbf{Commercial+RAG}} 
& GPT-3.5 + RAG               & 20.4 & 26.5 & 27.2 & 3.5 & \textbf{22.0} & 19.6 & 48.9 & 59.9 & 60.2 & 30.4 & 60.3 & 51.3 & 1.42 & 0.88 & 1.07 & 0.77 & 1.04 & 1.05 \\
& GPT-4o + RAG                & \textbf{29.3} & \textbf{27.6} & \textbf{28.9} & \textbf{12.7} & 21.2 & \textbf{23.9} & \textbf{52.2} & \textbf{60.7} & \textbf{61.9} & \textbf{37.1} & \textbf{61.5} & \textbf{54.2} & \textbf{1.46} & \textbf{1.20} & \textbf{1.46} & \textbf{1.30} & \textbf{1.19} & \textbf{1.33} \\
\bottomrule
\end{tabular}
}
\label{tab:qa_main}
\end{table*}

\begin{table*}[ht]
\centering
\caption{Performance of Mistral-24B under different RAG configurations. Answer prediction includes F1 Score, BERTScore, and LLM-as-Judge; retrieval accuracy includes R@k and NDCG@k.}
\resizebox{0.75\linewidth}{!}{
\begin{tabular}{ccccccc}
\toprule
\multirow{2}{*}{\textbf{Retrieval Granularity}} & 
\multirow{2}{*}{\textbf{Top-k}} & 
\multicolumn{3}{c}{\textbf{Answer Prediction}} & 
\multicolumn{2}{c}{\textbf{Retrieval Accuracy}} \\
\cmidrule(lr){3-5} \cmidrule(lr){6-7}
& & \textbf{F1 Score (\%)} $\uparrow$ & \textbf{BERTScore (\%)} $\uparrow$ & \textbf{LLM-as-Judge (0-2)} $\uparrow$ 
& \textbf{R@k (\%)} $\uparrow$ & \textbf{NDCG@k (\%)} $\uparrow$ \\
\midrule
\multirow{3}{*}{\textbf{Turn-level}} 
& \cellcolor{gray1}10  & 16.8 & 47.8 & 1.06 & 72.1 & 55.7 \\
& \cellcolor{gray2}20  & \textbf{20.3} & \textbf{50.1} & 1.08 & 86.4 & 60.7 \\
& \cellcolor{gray3}30  & 18.7 & 49.5 & \textbf{1.15} & \textbf{94.2} & \textbf{63.1} \\
\midrule
\multirow{3}{*}{\textbf{Round-level}} 
& \cellcolor{gray1}5   & 19.4 & 49.6 & 1.06 & 57.6 & 51.7 \\
& \cellcolor{gray2}10  & \textbf{20.4} & 50.1 & 1.15 & 70.8 & 56.6 \\
& \cellcolor{gray3}15  & 19.0 & \textbf{50.5} & \textbf{1.20} & \textbf{77.7} & \textbf{59.1} \\
\midrule
\multirow{3}{*}{\textbf{Session-level}} 
& \cellcolor{gray1}2  & \textbf{20.2} & \textbf{51.1} & 1.23 & 49.9 & 49.1 \\
& \cellcolor{gray2}4  & 18.8 & 50.4 & \textbf{1.27} & 65.0 & 55.9 \\
& \cellcolor{gray3}8  & 16.7 & 49.6 & 1.25 & \textbf{81.7} & \textbf{62.4} \\
\bottomrule
\end{tabular}
}
\label{tab:rag_result}
\end{table*}

\begin{table}[th]
\centering
\caption{Overall QA performance of Mistral models under different input context lengths. }
\resizebox{0.9\linewidth}{!}{
\begin{tabular}{c >{\centering\arraybackslash}p{1.2cm} ccc}
\toprule
\textbf{Model} & \textbf{Context} & \textbf{F1 Score} $\uparrow$ & \textbf{BERTScore} $\uparrow$ & \textbf{LLM-as-Judge} $\uparrow$ \\
\midrule
\multirow{4}{*}{Mistral-8B}
& \cellcolor{gray1}2K   & \textbf{9.8}  & \textbf{42.1} & \textbf{0.56} \\
& \cellcolor{gray2}4K   & 7.5  & 40.3 & 0.55 \\
& \cellcolor{gray3}8K   & 7.7  & 38.9 & 0.55 \\
& \cellcolor{gray4}20K  & 5.7  & 38.4 & 0.42 \\
\midrule
\multirow{4}{*}{Mistral-24B}
& \cellcolor{gray1}2K   & 16.2 & 47.2 & 0.80 \\
& \cellcolor{gray2}4K   & 14.4 & 46.9 & 0.82 \\
& \cellcolor{gray3}8K   & \textbf{17.4} & \textbf{48.6} & \textbf{1.04} \\
& \cellcolor{gray4}20K  & 15.5 & 47.4 & 1.01 \\
\bottomrule
\end{tabular}
}
\label{tab:context_truncation}
\end{table}

\section{Experimental Results}

We conduct a comprehensive evaluation of baseline methods on ES-MemEval to assess their ability to maintain and leverage long-term memory in personalized emotional support conversations.

\begin{table*}[t]
\centering
\setlength{\tabcolsep}{4pt}
\caption{Evaluation results on the summarization task of ES-MemEval.}
\resizebox{0.7\linewidth}{!}{
\begin{tabular}{cccccccccc}
\toprule
\multirow{2}{*}{\textbf{Category}} & \multirow{2}{*}{\textbf{Model}} 
& \multicolumn{3}{c}{\textbf{ROUGE (\%)} $\uparrow$} 
& \multicolumn{3}{c}{\textbf{Event-based Metrics (\%)} $\uparrow$} 
& \multirow{2}{*}{\textbf{LLM Score (0-5)} $\uparrow$} \\
\cmidrule(lr){3-5} \cmidrule(lr){6-8}
& & ROUGE-1 & ROUGE-2 & ROUGE-L & Precision & Recall & F1 & \\
\midrule
\multirow{2}{*}{\textbf{Base}} 
& Mistral-8B         & 21.6 & 3.5  & 12.0 & 20.1 & 25.5 & 21.7 & 1.23 \\
& Phi-3-Medium       & \textbf{25.1} & \textbf{5.7}  & \textbf{13.9} & \textbf{27.6} & \textbf{45.2} & \textbf{33.2} & \textbf{1.91} \\
& Mistral-24B       & 19.8 & 5.1  & 10.9 & 24.1 & 32.0 & 26.8 & 1.45 \\
\midrule
\multirow{2}{*}{\textbf{Base+RAG}} 
& Mistral-8B + RAG     & 32.6          & 6.7           & 17.8          & 39.2          & 44.1          & 40.6          & 2.34 \\
& Phi-3-Medium + RAG   & 28.0          & 6.3           & 15.2          & 34.0          & 50.2          & 39.6          & 2.34 \\
& Mistral-24B + RAG    & \textbf{37.4} & \textbf{8.7}  & \textbf{21.0} & \textbf{45.6} & \textbf{53.0} & \textbf{48.1} & \textbf{2.79} \\
\midrule
\multirow{2}{*}{\textbf{Commercial}} 
& GPT-3.5-turbo               & \textbf{35.3} & 6.8  & \textbf{19.3} & 23.3 & 29.1 & 24.7 & 1.75 \\
& GPT-4o                      & 35.0 & \textbf{8.6}  & 19.1 & \textbf{34.0} & \textbf{48.5} & \textbf{38.8} & \textbf{2.36} \\
\midrule
\multirow{2}{*}{\textbf{Commercial+RAG}} 
& GPT-3.5-turbo + RAG           & 39.3 & 9.3  & 21.7 & 44.3 & 50.0 & 46.2 & \textbf{2.93} \\
& GPT-4o + RAG                  & \textbf{40.5} & \textbf{11.1} & \textbf{22.2} & \textbf{46.0} & \textbf{54.3} & \textbf{49.4} & \textbf{2.93} \\
\bottomrule
\end{tabular}
}
\label{tab:sum_eval}
\end{table*}

\subsection{Analysis of QA Performance}

We conducted three QA experiments on ES-MemEval: (1) benchmarking models across five core memory abilities, (2) analyzing the impact of RAG configurations on answer quality and retrieval accuracy, and (3) evaluating the performance of Mistral models under varying context lengths.

\subsubsection{Comparison Across Models} 

As shown in Table~\ref{tab:qa_main}, RAG consistently drives performance gains across both open-source and commercial models. 
For instance, Mistral-24B with RAG exhibits notable improvement, with its overall F1 score rising from 15.5 to 18.8, BERTScore from 47.4 to 50.4, and LLM-as-Judge Score from 1.01 to 1.27, indicating that RAG enhances model robustness and effectiveness in ultra-long dialogues.
The gains are particularly pronounced for smaller models such as Mistral-8B, whose LLM Score increased by 0.43, highlighting that RAG is especially beneficial for models with limited capacity.
Despite these improvements, RAG performance remains uneven across different capabilities. 
Specifically, model performance in user modeling and temporal reasoning remains suboptimal, as F1 scores seldom exceed 20.0 even under RAG augmentation. 
Abstention varies sharply—RAG improves performance for open-source models, but reduces it for commercial ones.
GPT-4o is most affected, with its LLM Score declining from 1.67 to 1.30, suggesting that the added retrieved content may encourage overconfident responses.
Overall, RAG enhances factual recall but does not uniformly benefit all capabilities, underscoring the persistent difficulty of long-term user modeling.

\subsubsection{Impact of Retrieval Configurations}

We further investigate RAG configurations by varying both memory granularity and retrieval size $k$, as shown in Table~\ref{tab:rag_result}.
Table~\ref{tab:rag_result} presents the performance of Mistral-24B under three retrieval granularities: turn-level, round-level, and session-level. 
Among these, the session-level configuration, which retrieves entire sessions as memory units, achieves the highest performance, with an LLM-as-Judge score of 1.27 at $k=4$, surpassing the best round-level (1.20) and turn-level (1.15) settings. 
This finding indicates that session-level retrieval is more effective for long-term emotional support scenarios, where relevant information is sparsely distributed and only becomes meaningful when aggregated across multiple conversational turns.
Regarding retrieval accuracy, Recall@k consistently improves as $k$ increases, with values exceeding 75\% across all granularities.
However, NDCG@k remains moderate, with peak values ranging from 59\% to 63\%. 
These results suggest that while the retrieved units cover the majority of relevant content, their ranking quality is suboptimal, which may limit the effectiveness of downstream reasoning.

\subsubsection{Effect of Context Length}
Table~\ref{tab:context_truncation} evaluates Mistral-8B and Mistral-24B under varying context lengths. 
In the benchmark, each user’s full dialogue history spans 11K–19K tokens, meaning that context windows shorter than 20K require truncation.
Without RAG, Mistral-8B achieves optimal performance at 2K tokens, while Mistral-24B peaks at 8K.
These results indicate that although both models nominally support up to 128K tokens, their effective performance substantially deteriorates as input context length increases.
This degradation is particularly pronounced for the smaller 8B model, whose performance improves markedly when the context window is reduced, whereas Mistral-24B remains more robust, sustaining good performance at 20K tokens. 
Consistent downward trends across F1, BERTScore, and LLM-as-Judge further validate these findings.
These results highlight the limitations of long-context processing for small- to medium-scale models. 
Based on the findings in Table~\ref{tab:qa_main}, RAG can be considered as an effective approach to mitigate these limitations in dialogue system design.

\subsection{Analysis of Summarization}

As shown in Table~\ref{tab:sum_eval}, RAG substantially improves summarization performance for both open-source and commercial models. 
Among open-source systems, Mistral-24B + RAG achieves the highest performance, with ROUGE-L increasing from 10.9 to 21.0, event-level F1 rising from 26.8 to 48.1, and the LLM Score improving from 1.45 to 2.79. 
These gains not only narrow the gap with commercial systems but also enable Mistral-24B + RAG to surpass GPT-3.5-turbo + RAG in event-level evaluations.
Similar trends are observed in smaller open-source models such as Mistral-8B and Phi-3-Medium, although the improvements are less pronounced than those of Mistral-24B. 
On the commercial side, GPT-3.5-turbo + RAG and GPT-4o + RAG deliver the strongest overall performance, both attaining an LLM Score of 2.93, with GPT-4o + RAG achieving the highest event-level F1 (49.4). 
Overall, these results demonstrate that RAG markedly enhances models’ user modeling and temporal reasoning capabilities, enabling both open-source and commercial models to generate more coherent and information-rich summaries.

\begin{table}[t]
\centering
\caption{Observation-based evaluation results on the dialogue generation task of ES-MemEval. 
Metrics measure alignment with seeker observations.}
\resizebox{0.85\linewidth}{!}{
\begin{tabular}{cccccc}
\toprule
\textbf{Memory Setting} & \textbf{Model} & \textbf{Recall} $\uparrow$ & \textbf{Weighted Score} $\uparrow$ \\
\midrule
\multirow{5}{*}{No-Mem.} 
& Mistral-8B        & \textbf{0.28} & 0.28 \\
& Phi-3-Medium      & 0.25 & 0.26 \\
& Mistral-24B       & 0.20 & 0.20 \\
& GPT-3.5-turbo     & 0.26 & \textbf{0.30} \\
& GPT-4o            & 0.23 & 0.29 \\
\midrule
\multirow{5}{*}{Full-Hist.} 
& Mistral-8B        & 0.31 & 0.35 \\
& Phi-3-Medium      & 0.27 & 0.27 \\
& Mistral-24B       & 0.33 & 0.33 \\
& GPT-3.5-turbo     & 0.31 & 0.34 \\
& GPT-4o            & \textbf{0.35} & \textbf{0.36} \\
\midrule
\multirow{5}{*}{RAG} 
& Mistral-8B        & 0.34 & 0.40 \\
& Phi-3-Medium      & 0.29 & 0.32 \\
& Mistral-24B       & 0.35 & 0.41 \\
& GPT-3.5-turbo     & 0.37 & 0.44 \\
& GPT-4o            & \textbf{0.38} & \textbf{0.45} \\
\bottomrule
\end{tabular}
}
\label{tab:gen_obs}
\end{table}

\begin{table}[t]
\centering
\caption{LLM-as-Judge evaluation results on the dialogue generation task of ES-MemEval. 
Scores (1–5) are rated by GPT-4o on long-term memory (\textit{LT-Mem.}), personalization (\textit{Pers.}), and emotional support (\textit{ES}).}
\resizebox{0.85\linewidth}{!}{
\begin{tabular}{cccccc}
\toprule
\textbf{Memory Setting} & \textbf{Model} & \textbf{LT-Mem.} $\uparrow$ & \textbf{Pers.} $\uparrow$ & \textbf{ES} $\uparrow$ \\
\midrule
\multirow{5}{*}{No-Mem.} 
& Mistral-8B        & 2.42 & 2.79 & 3.26 \\
& Phi-3-Medium      & 2.90 & 3.53 & 3.21 \\
& Mistral-24B       & 1.42 & 2.84 & 2.53 \\
& GPT-3.5-turbo     & \textbf{2.95} & \textbf{3.79} & \textbf{3.32} \\
& GPT-4o            & 1.84 & 3.32 & 2.79 \\
\midrule
\multirow{5}{*}{Full-Hist.} 
& Mistral-8B        & 4.53 & 4.58 & 4.58 \\
& Phi-3-Medium      & 3.68 & 4.05 & 3.90 \\
& Mistral-24B       & 4.37 & 4.42 & 4.32 \\
& GPT-3.5-turbo     & 4.68 & 4.74 & 4.68 \\
& GPT-4o            & \textbf{5.00} & \textbf{5.00} & \textbf{5.00} \\
\midrule
\multirow{5}{*}{RAG} 
& Mistral-8B        & 4.63 & 4.74 & 4.74 \\
& Phi-3-Medium      & 4.37 & 4.58 & 4.42 \\
& Mistral-24B       & 4.90 & 4.90 & 4.90 \\
& GPT-3.5-turbo     & 4.63 & 4.74 & 4.68 \\
& GPT-4o            & \textbf{5.00} & \textbf{5.00} & \textbf{5.00} \\
\bottomrule
\end{tabular}
}
\label{tab:gen_llm}
\end{table}

\subsection{Analysis of Dialogue Generation}

\subsubsection{Observation-based Metrics}

As shown in Table \ref{tab:gen_obs}, under the No-Mem condition, scores remain low and any higher values in this setting largely reflect hallucinated user experiences rather than genuine memory use. 
By contrast, introducing Full-Hist or RAG substantially improved Recall and Weighted Score across all models—for example, Mistral-24B increased from 0.20 to 0.33—demonstrating that explicit histories provide more reliable representations of user states and enhance alignment with user observations.
Moreover, RAG variants consistently outperformed Full-Hist (e.g., Mistral-24B achieved 0.41 in Weighted Score versus 0.33 for Full-Hist), suggesting that external retrieval mechanisms help models explicitly leverage user-relevant information, thereby improving personalization and factual consistency.

\subsubsection{LLM Ratings}

Table \ref{tab:gen_llm} presents GPT-4o’s ratings of overall dialogue quality. 
Under No-Mem conditions, some models (e.g., GPT-3.5-turbo, Phi-3-Medium) still scored relatively high on LT-Mem., a result largely driven by their tendency to hallucinate plausible user experiences when lacking explicit memory access.
These results highlight the necessity of explicit long-term history: without it, models risk inconsistent personalization and compromised credibility. 
Compared to the No-Mem condition, access to long-term history markedly boosted long-term memory, personalization, and emotional support scores. 
Full-Hist and RAG achieved comparable performance, with RAG slightly better in some cases, indicating that retrieval can provide support comparable to full history while reducing context length. 
A key finding is the strong correlation between Pers. and LT-Mem., indicating that effective personalization depends on accurate memory recall. 
In contrast, ES scores are less sensitive to memory, suggesting that emotional support can partly rely on general strategies.

\section{Discussion and Future Directions}

Our ES-MemEval experiments reveal six key insights into long-term personalized emotional support dialogues. 
\textbf{(1) Long-term memory is essential}: without explicit histories, models may hallucinate user experiences, undermining reliability and personalization. 
\textbf{(2) RAG is effective but limited}: retrieval improves factual consistency and alignment with user observations, yet modeling nuanced temporal dynamics remains challenging, motivating retrieval-aware calibration. 
\textbf{(3) Memory drives personalization}: personalization strongly depends on long-term memory, whereas emotional support can partly rely on general strategies with limited memory. 
\textbf{(4) Retrieval configuration matters}: session-level retrieval better captures sparse and evolving user signals, though redundancy remains a concern. 
\textbf{(5) Long-context limits persist}: smaller models degrade with extended contexts, highlighting the need for memory–retrieval integration. 
\textbf{(6) RAG bridges system gaps}: RAG narrows the performance gap between open-source and commercial models by improving personalization and memory alignment. 
Together, these findings point to retrieval-aware calibration, adaptive memory granularity, and hybrid memory–retrieval designs for sustained personalized dialogue.

\section{Conclusion}

This paper presents the first benchmark study of long-term memory in personalized emotional support scenarios, a gap overlooked by prior long-term dialogue evaluations. We introduce EvoEmo, a dataset comprising 18 user trajectories with evolving user states across multiple sessions. Building on this resource, we propose ES-MemEval, a benchmark designed to systematically evaluate personalized dialogue agents' memory capabilities—including information extraction, temporal reasoning, conflict detection, abstention, and user modeling—through question answering, summarization, and dialogue generation tasks. Extensive experiments on open-source long-context, commercial, and retrieval-augmented models provide empirical insights into their ability to maintain and leverage long-term memory for personalization, highlighting the impact of factors such as memory granularity and retrieval strategies. Collectively, these contributions establish a high-quality, empirically grounded benchmark that facilitates the development of reliable, user-centered dialogue systems in complex, long-term settings.

\begin{acks}
This work was supported in part by the New Generation Artificial Intelligence-National Science and Technology Major Project under Grant 2025ZD0123701, in part by the National Natural Science Foundation of China under Grant 62476202 and 62272343, and in part by the Fundamental Research Funds for the Central Universities.
\end{acks}

\bibliographystyle{ACM-Reference-Format}
\bibliography{Bibtex/intro,Bibtex/method,Bibtex/related}

\appendix

\section{Limitations}

While this work introduces a novel benchmark and dataset for long-term emotional support conversations, several limitations remain.

First, the EvoEmo dataset is synthetic, generated with GPT assistance and refined through human review. This approach was chosen to mitigate ethical concerns and the substantial time cost of collecting large-scale, real-world long-term emotional support dialogues. Similar methods have been increasingly adopted as practical alternatives to labor-intensive manual curation \cite{r-10, intro-7, m-15, m-16}. Although user profiles and timelines were derived from real data and session consistency was manually verified, the dataset may still diverge from real-world conversational dynamics.

Second, while EvoEmo represents a substantial effort to model longitudinal user states, its overall size remains relatively small compared to large-scale general dialogue corpora. Nevertheless, due to its focus on long-term, multi-session interactions, the dataset's scale is comparable to, or even surpasses, that of existing domain benchmarks, such as LOCOMO (272 sessions) and MemoryBank (150 sessions with an average of 7.6 turns). Future work will prioritize increasing the number of users and sessions to enhance the dataset’s representativeness and complexity.

\begin{figure*}[t]
  \centering
  \subfloat[Topic distribution of dialogues
 ]{\includegraphics[width=0.33\linewidth]{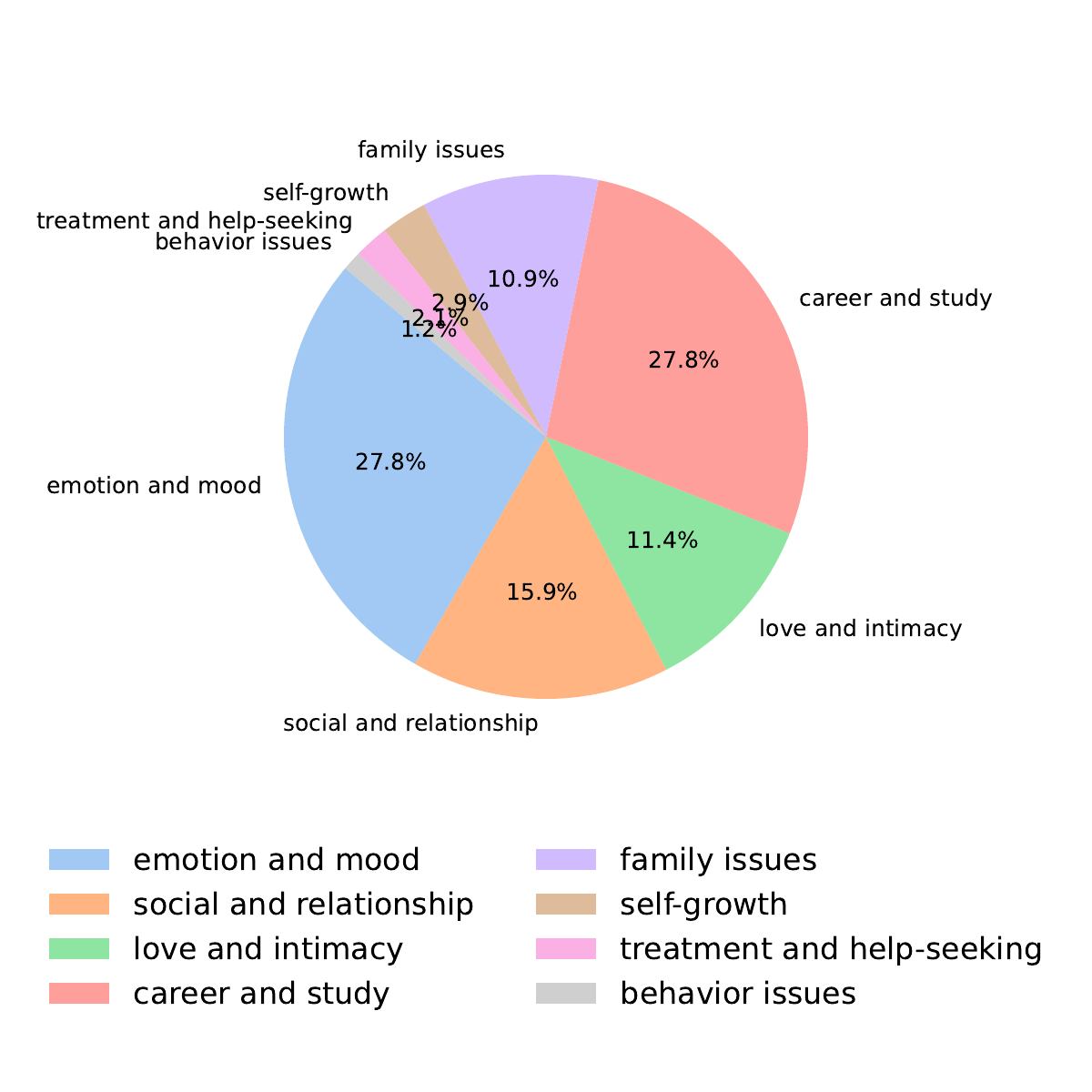}}
  \subfloat[Distribution of QA question types]{\includegraphics[width=0.33\linewidth]{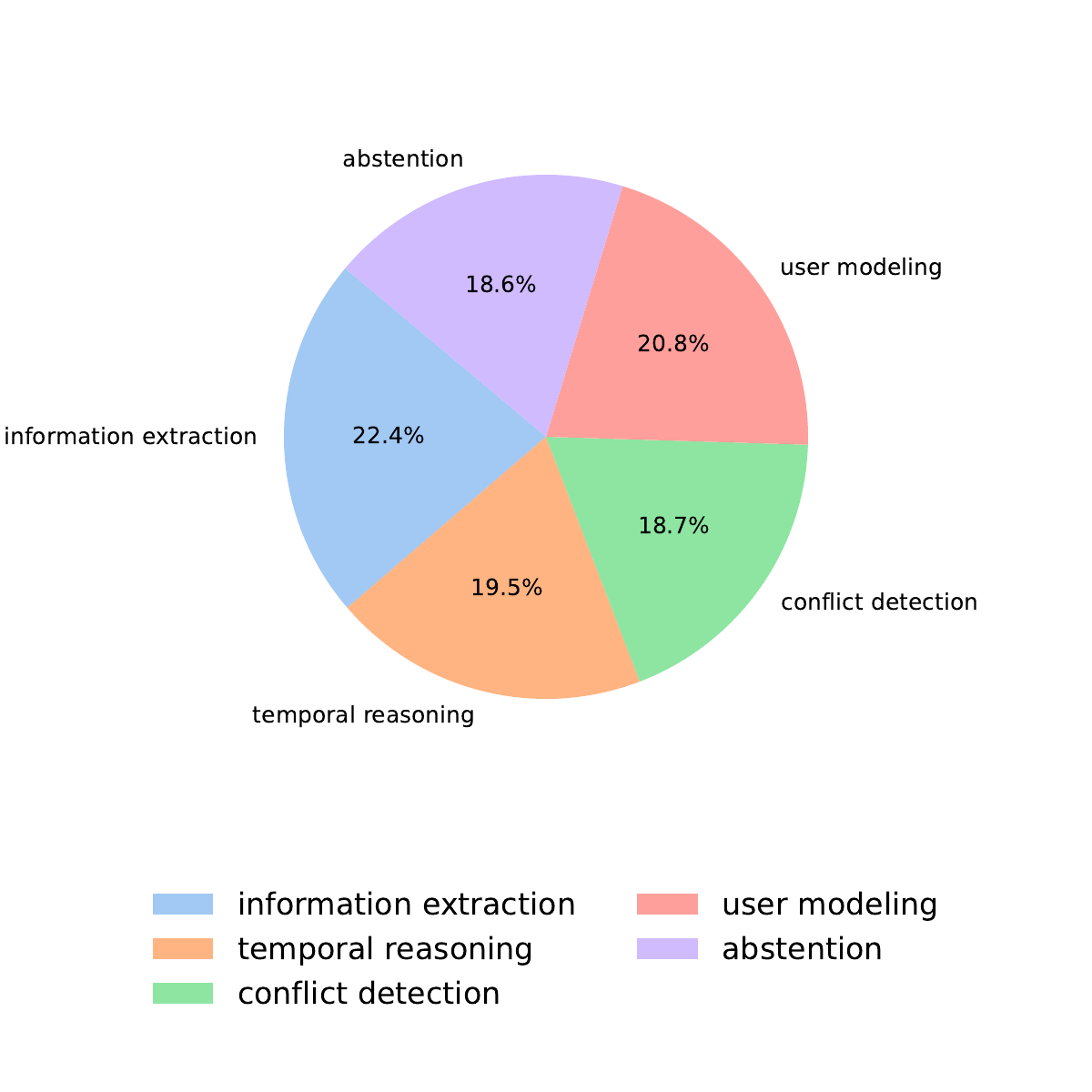}}
  \subfloat[Distribution of summarization question types]{\includegraphics[width=0.33\linewidth]{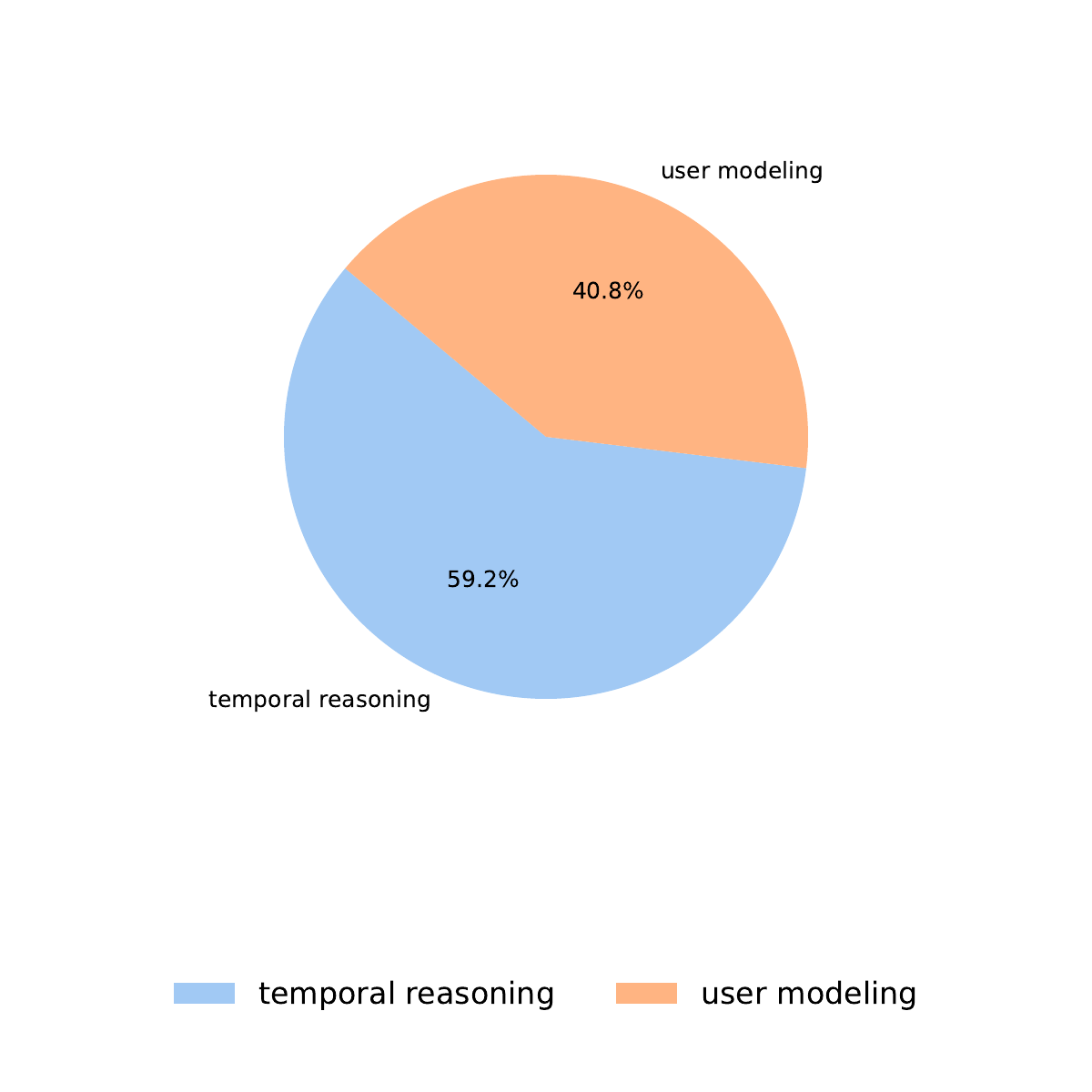}}
  \caption{Distributions of dialogue topics in EvoEmo and task types in ES-MemEval, covering QA and summarization.}
  \Description{Distributions of dialogue topics in EvoEmo and task types in ES-MemEval, covering QA and summarization.}
  \label{fig:task-distribution}
\end{figure*}

Third, as shown in Figure \ref{fig:task-distribution}, the dataset encompasses a diverse set of eight dialogue topic categories; however, the distribution is imbalanced and does not account for cross-cultural diversity. Categories such as \textit{self-growth} remain underrepresented, reflecting the skewed composition of the source dataset ESConv \cite{intro-8}. Expanding the dataset with additional sources is a key direction for improving scenario coverage and cultural representativeness.

Finally, our experiments primarily adopt common benchmark configurations and do not explore alternative retrieval algorithms (e.g., BM25 \cite{m-19}, DRAGON \cite{m-20}) or finer-grained memory units (e.g., user observations \cite{intro-7}, dialogue summaries \cite{intro-4, m-18}, or compressed contexts \cite{m-17}). While these remain promising directions, the primary goal of this work is to establish a benchmark rather than to optimize specific dialogue models or retrieval strategies. Future research can build on this foundation by exploring tailored solutions in these areas.

\begin{figure}[ht]
  \centering
  \subfloat[QA Task]{\includegraphics[width=0.5\linewidth]{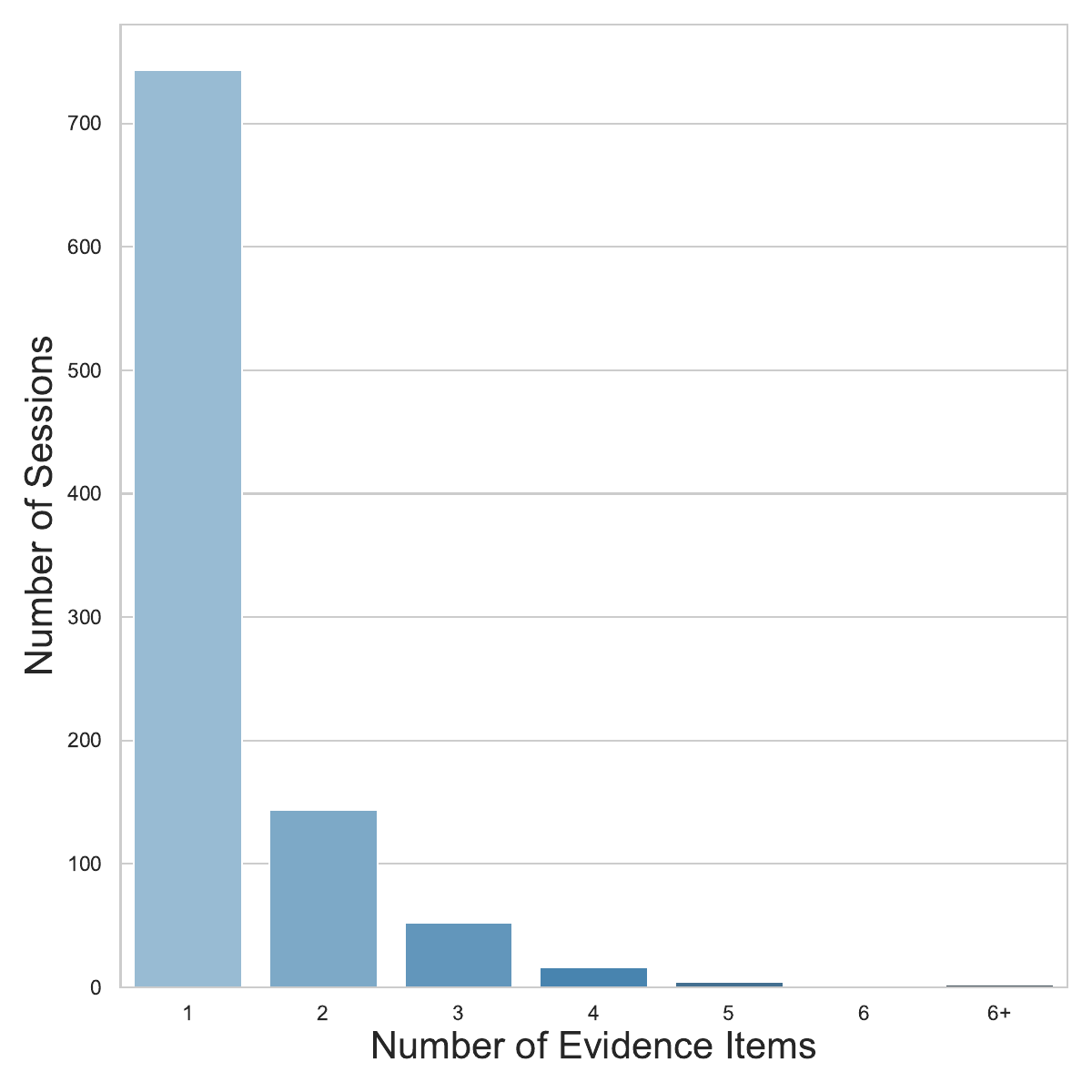}}
  \subfloat[Summarization Task]{\includegraphics[width=0.5\linewidth]{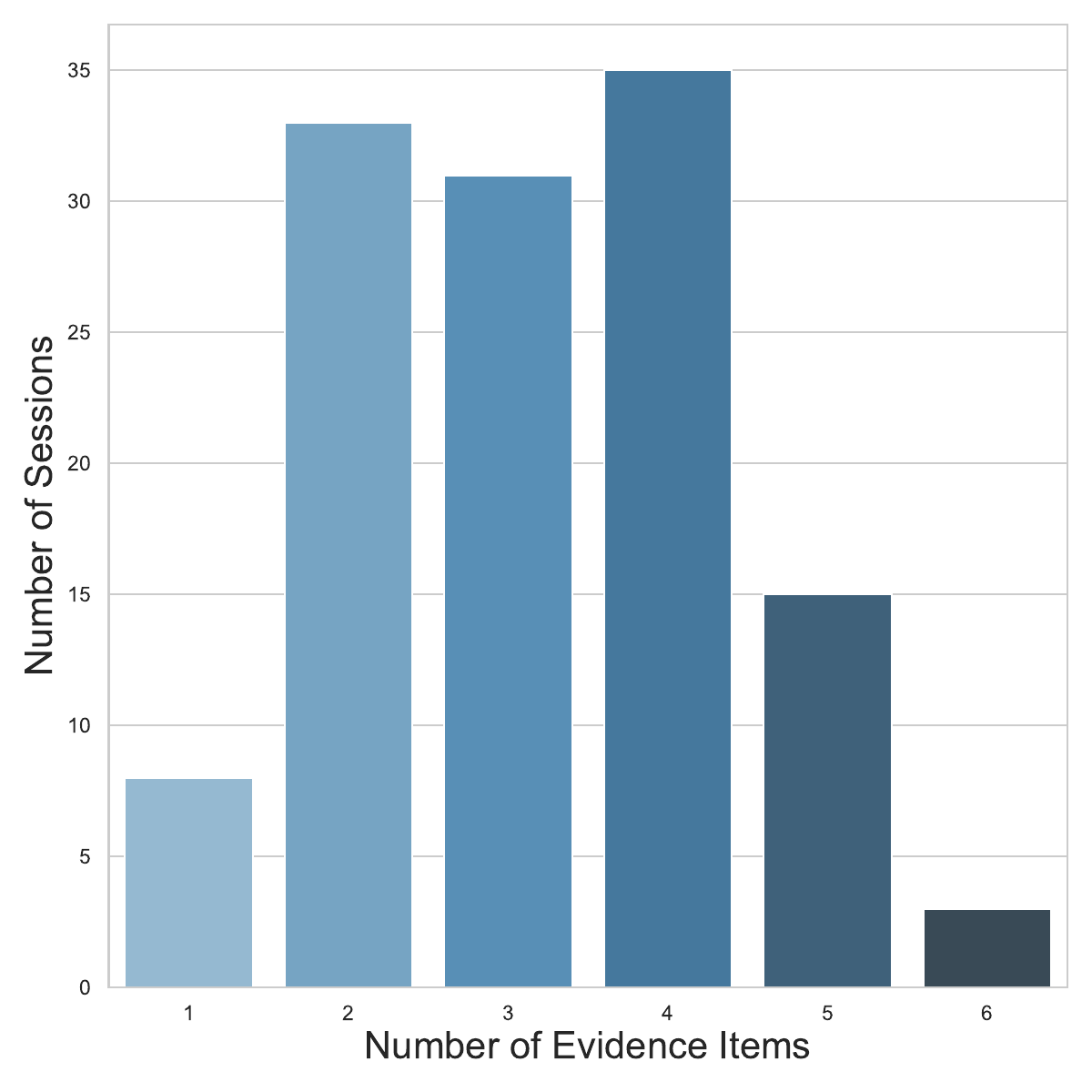}}
  \caption{Distribution of the number of evidence sessions in the QA and summarization tasks of ES-MemEval.}
  \label{fig:evidence-distribution}
  \Description{Distribution of the number of evidence sessions in the QA and summarization tasks of ES-MemEval.}
  
\end{figure}

\begin{figure}[thb]
  \centering
\includegraphics[width=0.95\linewidth]{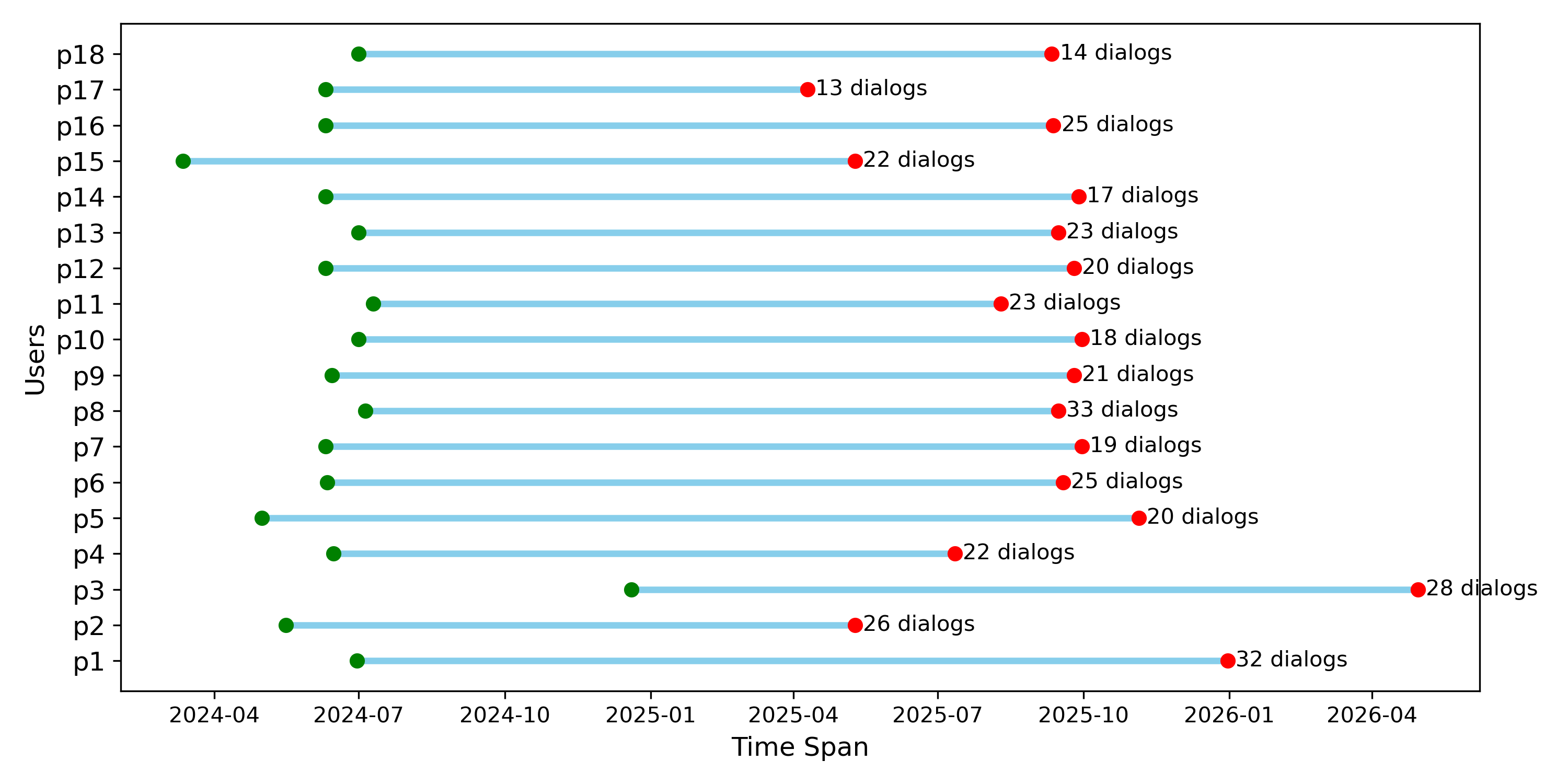}
  \caption{User-level timelines showing the span of dialogue histories from the first to the last recorded interaction.}
  \label{fig:time_span} 
  \Description{User-level timelines}
\end{figure}

\section{Ethics Statement}

This work introduces the EvoEmo dataset and the derived ES-MemEval benchmark, designed to evaluate conversational agents in long-term emotional support settings. To avoid the ethical and privacy risks of using real counseling or mental health data, we employed a synthetic pipeline in which GPT-generated dialogues were refined through multi-stage human review by trained annotators, who were fairly compensated. The dataset contains no real user conversations, eliminating risks of disclosing personal information, and was carefully audited for coherence, plausibility, and safety.

As ES-MemEval simulates sensitive scenarios, we caution that models—especially under memory-free conditions—may hallucinate user experiences or fabricate inconsistent histories. Such risks highlight the need for professional oversight, and ES-MemEval is released strictly for research purposes, not for real-world counseling or clinical deployment. We aim to advance research on long-term personalization while maintaining rigorous ethical standards of privacy, safety, and responsible use.

\section{Dataset}

\subsection{Annotator Details}

We recruited ten volunteers to assist with dataset construction. 
Two focused on refining and verifying event timeline expansions. 
Six were responsible for reviewing and correcting QA test samples and chat data. 
The remaining two contributed to the review of summarization and dialogue generation test samples. 
All volunteers were graduate students proficient in English, fully briefed on the task objectives and dataset annotation standards. 
They were compensated at a rate of 8 USD per hour.

\subsection{More Dataset Statistics \label{app:statistics}}

Figure~\ref{fig:task-distribution}(a) shows the topic distribution of EvoEmo, where dialogues mainly focus on \textit{emotion and mood} and \textit{career and study}, followed by \textit{social and relationship} and \textit{love and intimacy}. Additional categories, including \textit{family issues}, \textit{self-growth}, \textit{treatment and help-seeking}, and \textit{behavior issues}, are also covered, indicating diverse user states and interaction contexts despite varying frequencies.
Figures~\ref{fig:task-distribution}(b) and \ref{fig:task-distribution}(c) present the distribution of question types in the QA and summarization tasks of ES-MemEval. 
The QA benchmark covers five types in a relatively balanced manner, including information extraction, temporal reasoning, conflict detection, user modeling, and abstention). In contrast, the summarization task focuses more heavily on temporal reasoning and user modeling, thereby emphasizing these two more complex and challenging capabilities. 


Figures~\ref{fig:evidence-distribution}(a) and \ref{fig:evidence-distribution}(b) further analyze the evidence requirements. QA answers primarily rely on multiple utterances within a single session, highlighting the difficulty of within-session reasoning, whereas summarization often requires aggregating information across multiple sessions, strengthening the evaluation of cross-session integration and user trajectory modeling. Overall, these statistics indicate that the benchmark balances intra-session reasoning and cross-session user modeling.

In addition, EvoEmo demonstrates long-term engagement, as user dialogue histories last an average of 448 days (median 458, ranging from 304 to 553 days) with approximately 22 sessions per user. Figure~\ref{fig:time_span} illustrates these spans with user-level timelines, indicating that the corpus captures extended emotional support interactions over several months to years.

\begin{figure}[th]
\small 
\renewcommand{\baselinestretch}{0.9} 
\begin{tcolorbox}[title=QA Evaluation Prompt, center title]
\textbf{Task Description:}
You are an impartial evaluator. Your task is to score a model's answer to a given question against a gold (reference) answer.

\textbf{Scoring Criteria:} 
2: Completely correct and contextually accurate; 1: Partially correct but incomplete, vague, or missing key information; 0: Completely wrong or irrelevant.

\textbf{Input Format:} 
Question: \texttt{\{question\}}, Gold Answer: \texttt{\{gold\}}, Model Answer: \texttt{\{pred\}}

\textbf{Output Instructions:} 
Output only one line in the format: \texttt{Score: X},
where \texttt{X} is 0, 1, or 2. No other text or explanations should be generated.

\end{tcolorbox}







\begin{tcolorbox}[title=Summarization Evaluation Prompt (Compressed), center title]

Evaluate a model-generated summary against a human-written reference in long-term emotional support dialogues.  

\textbf{Event Definition:}  
An \texttt{event} is a distinct state change, emotional shift, or decision explicitly mentioned in the summary, for example, ``\textit{considered quitting a job}''.

\textbf{Evaluation Steps:}  
1. Extract \texttt{reference events} and \texttt{generated events}.  
2. Identify \texttt{recalled events}, i.e., the semantic intersection of \texttt{reference events} and \texttt{generated events}.
3. Count events: $\#$reference (\texttt{reference events}), $\#$generated (\texttt{generated events}), and $\#$recalled (semantic overlaps).

\textbf{Scoring (0–5):}  
Score based on event recall, hallucinations, and consistency; lower scores for omissions or inconsistencies.

\textbf{Output Format:}
JSON with score (0–5), event lists and counts, plus a brief justification.

\end{tcolorbox}

\begin{tcolorbox}[title=Dialogue Evaluation Prompt (Compressed), center title]

Evaluate the supporter in long-term emotional support dialogues on a 1–5 scale:

\textbf{Memory:} Accurately recall and integrate the Seeker’s past experiences. Lower scores for omissions, incomplete references, inaccuracies, or vague mentions.  

\textbf{Personalization:} Responses should be tailored to the Seeker’s experiences, personality, and preferences, offering concrete guidance. Generic or templated responses must not exceed 3.  

\textbf{Emotional Support:} Provide empathy, reassurance, or encouragement explicitly grounded in the Seeker’s experiences and emotions. Limited, generic, or cold responses lower the score.  

\textbf{Scoring Instructions:}  
Provide an integer 1–5 for each criterion based on accurate memory, tailored personalization, and grounded emotional support; generic/template responses must not exceed 3.

\end{tcolorbox}

\caption{Compressed LLM-as-Judge Prompts in ES-MemEval.}
\label{fig:eval_prompt}
\Description{Compressed LLM-as-Judge Prompts for Benchmarking.}
\end{figure}

\section{Experimental Setup}

\subsection{Evaluation Metrics \label{sec:app_metric}}

We developed a suite of LLM-as-judge prompts to systematically evaluate three tasks: question answering, summarization, and dialogue generation. 
Each task is paired with a dedicated evaluation prompt tailored to its specific requirements. 
For brevity, condensed versions of the summarization and dialogue generation prompts are presented in Figure~\ref{fig:eval_prompt}, while the full set of prompts will be released in our code repository to ensure transparency, reproducibility, and future benchmarking.

\begin{table}[bht]
\centering
\caption{
Consistency between human annotators and \textit{LLM-as-Judge} across QA, summarization (Sum.), and dialogue generation (DG) tasks. Reported metrics include Weighted Cohen’s Kappa ($\kappa$), Spearman correlation ($\rho$), and Mean Absolute Difference (MAD). For DG, human ratings are heavily skewed toward the maximum, producing artificially low $\kappa$ and $\rho$; MAD and exact agreement better reflect alignment.
}
\resizebox{\linewidth}{!}{
\begin{tabular}{lccccccccc}
\toprule
\multirow{2}{*}{\textbf{Model}} & 
\multicolumn{3}{c}{\textbf{QA}} & 
\multicolumn{3}{c}{\textbf{Sum.}} & 
\multicolumn{3}{c}{\textbf{DG}} \\
\cmidrule(lr){2-4} \cmidrule(lr){5-7} \cmidrule(lr){8-10}
& $\kappa \uparrow$ & $\rho \uparrow$ & MAD $\downarrow$ 
& $\kappa \uparrow$ & $\rho \uparrow$ & MAD $\downarrow$ 
& $\kappa \uparrow$ & $\rho \uparrow$ & MAD $\downarrow$ \\
\midrule
Mistral-24B+full     & 0.72 & 0.73 & 0.25 & 0.78 & 0.66 & 0.44 & 0.57 & 0.61 & 0.40 \\
Mistral-24B+RAG & 0.69 & 0.71 & 0.26 & 0.60 & 0.66 & 0.50 & 0.19 & 0.20 & 0.17 \\
\bottomrule
\end{tabular}}
\begin{tablenotes}
\small
\item \textit{Note.} Despite moderate or low Kappa and Spearman values for DG due to ceiling effects, MAD (0.40 for Mistral-24B+full, 0.17 for Mistral-24B+RAG) and exact agreement rates (70\% and 86.7\%) indicate that \textit{LLM-as-Judge} captures meaningful distinctions and overall alignment with human judgments.
\end{tablenotes}
\label{tab:judge_consistency}
\end{table}

For the observation-based protocol in the dialogue generation task, we designed an evaluation method that directly measures whether system responses utilize user information. 
First, we constructed a candidate set of user observations, defined as objective descriptions of user states, experiences, or contextual facts, extracted from scenario-related conversations. 
For each dialogue turn, Mistral-24B performed two tasks: (i) scoring the relevance of each observation to the user’s current input on a discrete scale of {0, 0.5, 1}, and (ii) assessing whether the system’s response explicitly or implicitly leveraged any observations deemed relevant (i.e., with a score of 0.5 or 1).
From these annotations, we computed two metrics: (1) \textit{Observation Recall} – the proportion of fully relevant observations (score = 1) that were reflected in the system’s responses across the test scenarios.
(2) \textit{Weighted Accuracy} – an aggregate score that accounts for both fully and partially relevant observations, providing a finer-grained measure of how well the system incorporated available user information.


\subsection{Reliability Analysis of LLM-as-Judge Evaluations}



To assess the reliability of the \textit{LLM-as-Judge} protocol, we sampled 50 QA, 40 summarization, and 30 dialogue examples, comparing human evaluations of Mistral-24B+Full and Mistral-24B+RAG using Weighted Cohen’s Kappa \cite{m-23}, Spearman’s rank correlation \cite{m-21}, and Mean Absolute Difference (MAD) \cite{m-22}. As shown in Table~\ref{tab:judge_consistency}, QA and summarization exhibit strong agreement with human judgments (Kappa > 0.6, Spearman > 0.6, MAD < 0.5), indicating that \textit{LLM-as-Judge} reliably captures both overall trends and fine-grained distinctions for these tasks. For dialogue generation, agreement is moderate for the full model (Kappa 0.57, Spearman 0.61) and lower for the RAG variant (Kappa 0.19, Spearman 0.20), primarily due to ceiling effects in human ratings resulting from highly positive general model quality. However, MAD (0.40 and 0.17) and exact agreement (70\% and 86.7\%) suggest that \textit{LLM-as-Judge} still aligns reasonably with human assessments, capturing meaningful distinctions even in open-ended, personalized dialogue settings. These results support the general reliability of the protocol while highlighting task-dependent variations in agreement.

\end{document}